# Automatic Tip Detection of Surgical Instruments in Biportal Endoscopic Spine Surgery

Sue Min Cho[1,2], Young-Gon Kim[1], Jinhoon Jeong[1], Ho-jin Lee[3], Namkug Kim[1]

[1] Asan Medical Center
[2] Johns Hopkins University
[3] Chungnam National University School of Medicine

## Abstract

*Some endoscopic surgeries require a surgeon to hold the endoscope with one hand and the surgical instruments with the other hand to perform the actual surgery with correct vision. Recent technical advances in deep learning as well as in robotics can introduce robotics to these endoscopic surgeries. This can have numerous advantages by freeing one hand of the surgeon, which will allow the surgeon to use both hands and to use more intricate and sophisticated techniques. Recently, deep learning with convolutional neural network achieves state-of-the-art results in computer vision. Therefore, the aim of this study is to automatically detect the tip of the instrument, localize a point, and evaluate detection accuracy in biportal endoscopic spine surgery. The localized point could be used for the controller's inputs of robotic endoscopy in these types of endoscopic surgeries.*

## 1. Introduction

There are several endoscopic surgeries performed today including plastic surgery, panendoscopy, orthopedic surgery, endodontic surgery, endoscopic endonasal surgery, and endoscopic spinal surgery. In general, due to the nature of these surgeries, the surgeon must hold the endoscope with one hand (usually the left hand) and hold the surgical instruments to perform the desired surgery with the other hand (usually the right hand) [1] or an assistant must hold the endoscopy to preserve the surgeon's view, which could cause several challenges. Figure 1 shows an overview of biportal endoscopic spine surgery (BESS)[2, 3]. In case of the surgeon holding the scope, the surgeon performs the operation with only one hand, which could lead to longer operation time, considerable learning period [4, 5] and possibly lower surgical outcomes. Otherwise, in case of the assistant holding the scope, an additional assistant is required which leads to additional cost and communication problems between surgeons and operators.

Therefore, introducing a robotic endoscopy and vision intelligence to these types of surgical techniques can have numerous advantages by freeing one hand of the surgeon. It can allow the surgeon to use both hands for the surgery and focus solely on the surgery, which can significantly reduce surgery time. In addition, this can shorten the learning curve or the time to master the surgery. Since its approval in 2000, the use of robotics in surgery has expanded progressively [6]. Kim et al. reports that endoscopic treatment in spine surgery is extending to more advanced lesions and that it is likely for endoscopy to be an available option for all spine disorder treatments based on the current rate of development [7]. Thus, the introduction of robotics is imminent, and its vision intelligence is vital for the robot to control the endoscope for the surgeon.

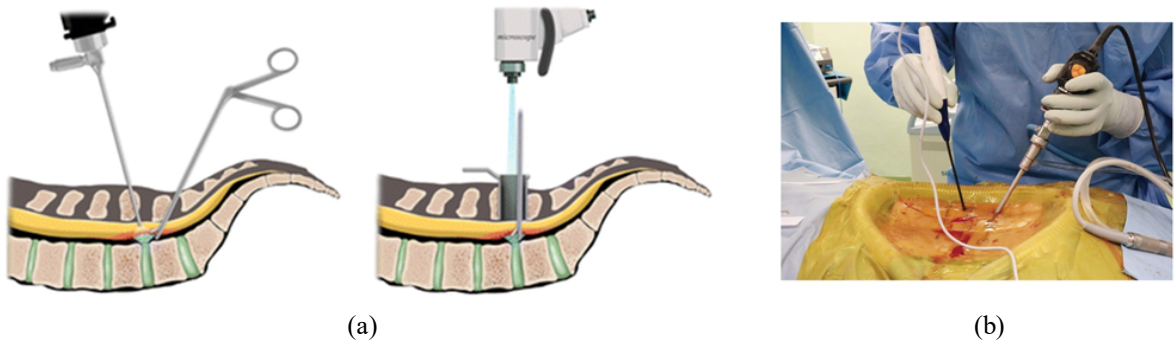

(a)          (b)

Figure 1: Overview of BESS. (a) Illustration of BESS (left) and uniportal endoscopic spine surgery (right), (b) Surgeon performing BESS.

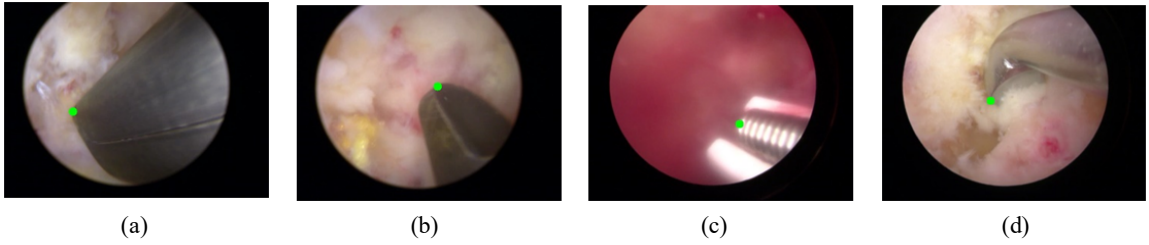

Figure 2: Example images from BESS videos showing large variations of shape, size, location of instrument, and corresponding tip annotations visualized as a green point with various instruments. (a) Kerrison punch, (b) pituitary rongeur, (c) radiofrequency probe, (d) curette.

For developing vision intelligence in biportal endoscopic spine surgery (BESS), this study aims to intelligently detect the tip of the instrument and localize a point in videos of BESS, which would be the point the surgeon needs to focus on during the surgery. The tip detection could be used for controller's inputs of robotic endoscopy in these endoscopic surgeries.

## 2. Related Works

Numerous approaches to track surgical instruments have been proposed in previous studies. In this section, we will present the related works to our study. They can be divided into two categories including traditional vision-based and deep learning methods.

### 2.1. Traditional vision-based methods

Since the advent of minimally invasive surgery such as robotized laparoscopy, numerous studies have aimed to find methods to detect and track surgical instruments. Traditional vision-based methods have been relied on hand-crafted features such as color and shape.

Doignon C. *et al.* used color properties to detect and localize grey regions of instruments in color images [8] and Wolf R. *et al.* used a geometric model of the instrument and the probabilistic condensation algorithm for 3D tracking of instruments [9]. Uecker D. *et al.* proposed an automated instrument tracking method based on automated image analysis and robotic visual servoeing in robot-assisted laparoscopic surgery [10].

Traditional vision-based methods are generally known to be reliable and inexpensive. However, the biggest limitation of these methods is that the features must be pre-defined and that it is hard to inference with the extracted features in real time. Also, it is known that these features cannot be extended ton difficult to extend them to messy and occluded images [11]. Features need to be fast and be robustly extracted. Thus, it is difficult to use these methods for real-time tracking of instruments

### 2.2. Deep learning methods

Deep learning methods in the field of computer vision have been drastically advancing and expanding due to the large amounts of data available and much improved computational power than before. A significant difference that deep learning techniques have compared to the traditional computer vision techniques is the automated direct extraction of optimal high-level features. Thus, deep learning is currently the dominant technique used for image recognition, and according to Litjens G. et al., deep learning boasts prevalence in every aspect of medical image analysis [12].

EndoNet [13] was the first convolutional neural net (CNN) based model for recognition tasks that included tool presence detection and phase recognition in laparoscopy. Following this work and with the advent of different and more accurate neural network architectures, various studies have been conducted regarding instrument tracking in surgical videos.

Deep neural networks for pose estimation of instruments have been proposed such as the paper of Du X. et al. which presents a model formed by a fully convolutional detection-regression network for articulated 2D pose estimation of multiple instruments [14]. Zhao Z. et al. presented automatic real-time tracking of surgical instruments based on spatial transformer network and spatio-temporal context [15]. Vardazaryan A. et al. and Nwoye et al. proposed weakly supervised learning for tool tracking, which enables localizing tools without explicit spatial annotations [16, 17]. Contextual detector of surgical tools based on recurrent convolutional neural network (RCNN) called LapTool-Net has recently been proposed as well [18].

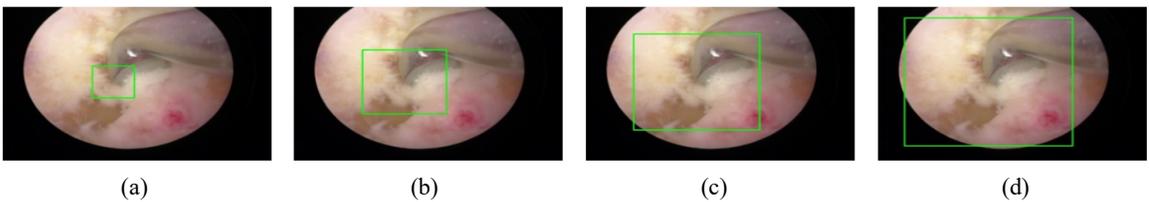

Figure 3: Examples of various margin sizes including (a) 50, (b) 100, (c) 150, (d) 200 pixels for tip detection, respectively.

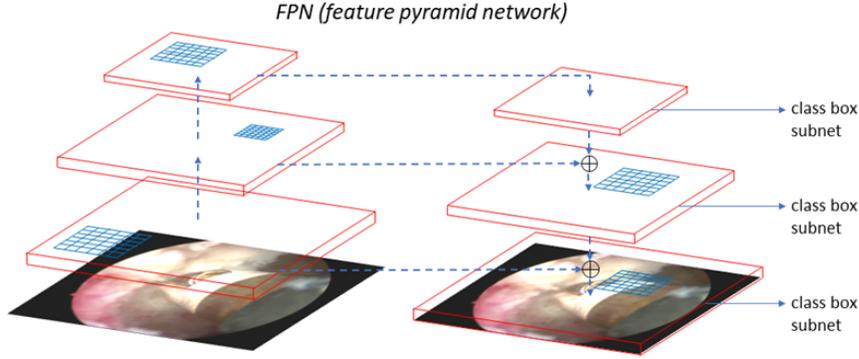

Figure 4: RetinaNet network architecture

The datasets of the previous studies have mostly been publicly available existing datasets such as m2cai16-tool, RMIT, EndoVis and Cholec80. There have been numerous studies on surgical instrument detection, tracking, and pose estimation in retinal microsurgery, cholecystectomy and laparoscopy. In our study, we are focusing on tip detection in BESS. We need a mechanism for BESS to follow and track the point that the surgeon needs to see. Not only has this domain of BESS not been investigated regarding instrument tracking, but also the method of automatic tip detection and point localization has not been explored before.

## 3. Dataset

The dataset consists of 9 BESS videos acquired from the department of orthopedic surgery of XX University of Medicine. The videos were recorded at 30 fps with a resolution of 640×480, and still images were captured from them for every 300 frames. The image was removed if no instrument was visible. As a result, the dataset contains 2310 frames annotated with the x, y coordinates of the tip by an expert.

The coordinates were labeled based on the guideline and definition of the tip of the instrument as follows. The tip of the instrument is selected to be the point of interest to be localized. There is a substantial number of moments when the instrument tip is covered by small disk fragments, bone spurs, other soft tissue, or bubbles. In these cases, the point is selected to be the nearest clearly visible point to the tip of the instrument. For the consistent usage of the term, these points are all inclusively referred to as the "tip" of the instrument. Some examples of the annotated data are shown in figure 2.

We randomly divided the dataset into training, validation, and test in 7:1:1 ratio. Because images in the same video should not be included in the three groups interchangeably, the groups were divided based on the videos. Consequently, training, validation, and test sets comprised of 1766 still images from 7 videos, 304 still images from 1 video, and 240 still images from 1 video, respectively.

## 4. Method

Inspired by the study of Kim Y.-G. et al. [19], we trained two detectors, RetinaNet and YOLOv2 with different margin sizes for the bounding box annotations to see if there is an optimal margin size for detecting the tip of the instrument and localizing the point. With the x, y coordinates of the instrument tip as the center of the bounding box, x1, y1, x2, y2 were calculated based on the margin size, M. The margin refers to the distance from the tip, and 50, 100, 150, and 200 pixels were used in the experiments. Because our target is the tip, M×M bounding box masks were made with the labeled tip coordinates as the center point. Figure 3 shows examples of various margin sizes: 50, 100, 150, and 200 pixels.

Due to the nature of the dataset, there is only one instrument and its corresponding tip per image. For this, we modified the algorithm to detect and output one bounding box that has the highest confidence out of the boxes that have confidences that are over 0.1 during inference. We set a threshold of 0.1 because when the highest confidence is lower than 0.1, the results are out of range (i.e. making a detection in the black region outside of the endoscopic view), so we determined to set a minimum threshold for the confidence level to consider the results with confidences lower that the threshold as false negative.

### 4.1. RetinaNet

In this experiment, RetinaNet [20] was implemented for instrument tip detection in BESS. RetinaNet is a simple dense one-stage detector that matches the accuracy of state-of-the-art two-stage object detectors. By introducing a focal loss, it embraces fast speed as well as high accuracy.

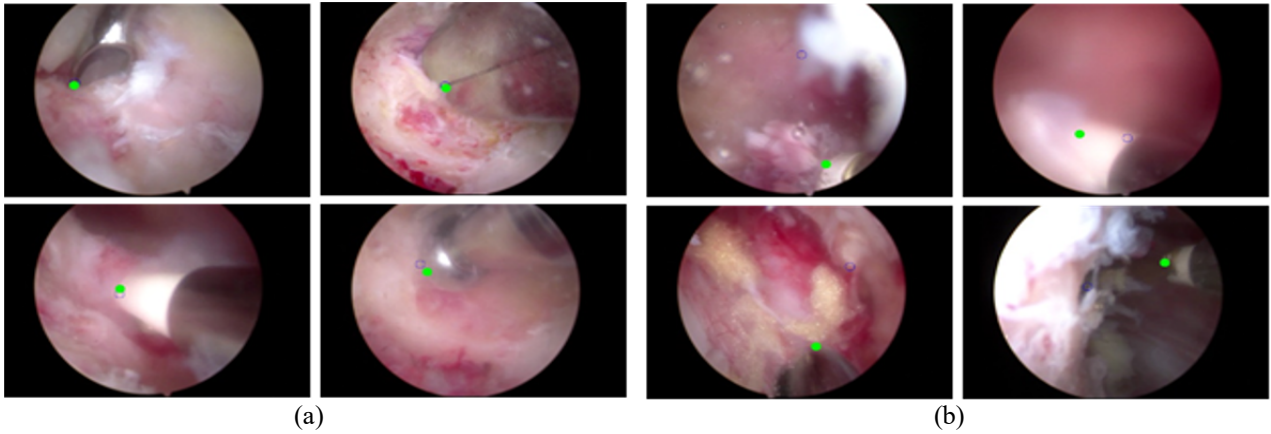

Figure 5: Result comparisons detected by RetinaNet based deep learning model for (a) excellent and good cases and (b) poor cases (ground truth point marked as green full circle and prediction point marked as blue empty circle).

It has ResNet and feature pyramid network as backbone for feature extraction, and two task-specific subnetworks for classification and bounding box regression. Figure 4 shows the architecture of RetinaNet.

The tip of the instrument was detected using RetinaNet with ImageNet pre-trained model based Resnet50. Data augmentation methods used were rotation (-0.1, 0.1 rad), translation (-0.1, 0.1), shear (-0.1, 0.1), and scaling (0.9, 1.0). The model was implemented in Keras 2.2.4 with Tensorflow 1.8.0 as backend and was trained with a Tesla P40 GPU. Smooth L1 loss was used for box regression and focal loss was used for classification. Adam optimizer with learning rate 0.0001 as used for region proposal and classification network. The learning rate was reduced on plateau.

**4.2. YOLOv2**

In this experiment, YOLOv2 [21] was implemented for instrument tip detection in BESS for comparison with RetinaNet. YOLOv2 is a real-time object detector that does not have any fully connected layers and uses anchor boxes to make predictions. It is known to achieve lower accuracy, but much faster speed compared to other state-of-the-art detectors. We wanted to investigate whether different networks have a different optimal margin size. We also wanted to measure the performance difference between RetinaNet and YOLOv2 regarding tip detection. The Darknet19 pre-trained model was used to fine-tune the network. The model was trained with a GeForce GTX 1080 GPU. Sum-squared error, composed of classification loss, localization loss, and confidence loss was used. Adam optimizer with learning rate 0.0001 was used for region proposal and classification network. The learning rate was reduced on plateau.

**4.3. Nine-fold cross-validation**

Then, we performed nine-fold cross-validation on the better performing network with its optimal margin size to validate its overall performance. The range of the number of images for the training, validation, and test sets were 1714-1942, 143-357, 143-357, respectively.

**5. Measurement**

Because our task is to detect the tip, we did not use the typical bounding boxes that are used in detection but used bounding boxes centered around the annotated tip coordinates with specific margin sizes. Thus, we used measurements that are different from the typical object detection measurements to find the optimal margin size and analyze the performance for each network. We calculated the midpoints with the predicted bounding box coordinates. With these predicted midpoints, we used the following method to measure the tip detection performance.

We calculated the recall, precision, $F1$-score with a fixed box size for both the ground truth tip coordinates and predicted midpoints, in which cases with Intersection over Union (IoU) score > 0.5 were considered "good" predictions. A fixed box size was determined in order to make fair comparison of the performance of models trained with different margin size bounding boxes by the 194 pixels in height and 192 pixels in width calculated from the dataset of 2,276 device bounding box masks.

**6. Results**

In this section, we will present the quantitative and qualitative results of our experiments.

| Margin size (pixel) | Recall | Precision | *F1*-score |
|---|---|---|---|
| 50 | 0.877 | 0.650 | 0.747 |
| 100 | 0.988 | 0.714 | 0.829 |
| **150** | **1.000** | **0.733** | **0.846** |
| 200 | 1.000 | 0.700 | 0.824 |

Table 1. RetinaNet result comparisons with various margin size in terms of recall, precision, and *F1*-score.

| Margin size (pixel) | Recall | Precision | *F1*-score |
|---|---|---|---|
| 50 | 0.738 | 0.797 | 0.766 |
| 100 | 0.750 | 0.810 | 0.779 |
| **150** | **0.864** | **0.808** | **0.835** |
| 200 | 0.885 | 0.775 | 0.826 |

Table 2. YOLO v2 result comparisons with various margin size in terms of recall, precision, and *F1*-score.

### 6.1. RetinaNet

For RetinaNet, the training time was 2.5 hours on average and inference time was 13.21 (±0.63) fps on average. Table 1 shows the results for this algorithm. Margin 150 was found to be the most optimal with recall of 1.000, precision of 0.733, and *F1*-score of 0.846. In this optimal case, there were 176 true positives, 64 false positives, and no false negatives.

Figure 5(a) shows examples of excellent and good cases. Visual analysis shows that the tip is well detected when the instrument and its tip is clearly visible and is positioned near the center. Figure 5(b) shows examples of poor cases. Visual analysis shows that the difficulty of detecting the tip increase when the instrument is far away, when the instrument is under the influence of shadow formed by tissue that is located closer to the camera lenses, when the image has extreme motion blur or occlusion.

### 6.2. YOLOv2

In case of YOLOv2, the training time was 3.5 hours on average and inference time was 71.43 fps on average. Table 2 shows the results for this algorithm. Margin 150 was found to be the most optimal with *F1*-score of 0.83. In this optimal setting, there were 172 true positives, 41 false positives, and 27 negatives that were detected. The mean distance calculated was 37.48±40.33 pixels.

In comparison between RetinaNet and YOLOv2, YOLOv2 has a faster inference time of 71.43 fps. In addition, we were able to see there was an approximately 2% difference in *F1*-score, and RetinaNet results have no false negatives whereas YOLOv2 results do, which means that RetinaNet performs better.

### 6.3. Nine-fold cross-validation

Table 3 shows the results of the 9-fold cross-validation. The optimal margin 150 of RetinaNet was used to cross-validate its overall robustness. The mean recall was 1.000 ± 0.000, precision 0.767 ± 0.0330, and f1-score 0.868 ±0.0223.

| Margin size (pixel) | Recall | Precision | *F1*-score |
|---|---|---|---|
| Mean ± SD | 1.000 ± 0.000 | 0.767 ± 0.033 | 0.868 ± 0.022 |

Table 3. Results of the nine-fold cross-validation.

### 6.4. Discussion and Conclusion

In this study, we have presented an automatic tip detection method for surgical instruments in endoscopic surgery, compared two state-of-the-art detection algorithms, RetinaNet and YOLOv2, and validated the overall robustness with cross-validation.

Through the optimal margin size experiments, we were able to see that there is an optimal margin size for the tip detection. However, this optimal margin size for the tip detection task may depend on this dataset. For future work, we would like to perform extra-validation to investigate this concern and determine the full robustness of the model.

When the bounding box is too small, the feature map of the tip gets smaller, which can lead to information shrinkage and loss. This can explain the lower performance results for smaller margin sizes shown in table 1 and 2. On the other hand, if we train the models with bounding boxes enlarged more than necessary, it is likely that the model will be trained with not only the information of the tip but also the information of many obstacles that can cause false positive results. Therefore, as seen in table 1 and 2, performance increases as the margin increases but saturates or decreases from at some point.

RetinaNet was concluded as the algorithm with better performance because it had a better *F1*-score. However, with its inference time being 13.21 (±0.63) fps, in order to make it real-time we must make the algorithm make an inference for every 2 or 3 frames. If a real-time detection is needed immediately, YOLOv2 will be an appropriate option. It is not only fast but also obtains results not too different from RetinaNet. It is shown in previous studies that YOLOv2 gains better accuracy when ResNet is used as its backbone architecture. For future work, we aim to investigate ways to increase the performance accuracy for YOLOv2.

In addition, due to the small number of videos, only 1 video was used for the validation set and 1 video for the test set. However, this was supplemented by the nine-fold cross-validation. Through the cross-validation of RetinaNet with the optimal margin size, we validated its detection performance with satisfactory results overall quantitatively.

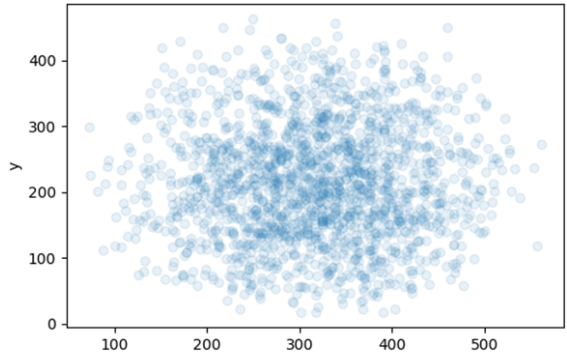

Figure 6: Scatterplot of tip coordinates of entire dataset.

There are some limitations of the dataset we used in this study. First, we only used still images. Because the end goal of this task is to be applied in a video, we aim to further this work to testing the models on videos as well. Second, we only used images in which an instrument is visible. For training, it makes sense to use images with the instruments, but it is slightly unrealistic for testing, because there are moments where the instrument is not present in actual BESS videos. Third, the dataset contained one instrument per image. This was inevitable in that BESS is currently performed with one instrument at a time. However, when two hand BESS becomes possible with the advent of an endoscopy robot, more than one tip will be present. For future work, we aim to modify and develop the automatic tip detection method for surgical instrument so that it can detect more than one tip at a time.

Based on the visual analysis of the excellent, good, and poor results, one of the main factors affecting the results seemed to be the location of the tip. It seemed that the tip was generally better detected when close to the center and vice versa. We drew a scatterplot with the x, y annotated tip coordinates of the entire dataset (2310 images from 9 videos) shown in figure 6. It can be visually seen that most of the coordinates are near the center. Also, the mean and median of x coordinates was 316.30 (±88.44) and 320 and the mean and median of y coordinates was 214.52 (±88.02) and 210. So, quantitatively it can also be concluded that most of the coordinates are near the center. We believe that increasing the number of images with tip on the periphery in the training set through data augmentation or data curation may increase the performance and make the model even more robust.

In conclusion, we presented an automatic tip detection method for surgical instruments in endoscopic surgery and validated its performance. This method can be extended to different types of endoscopy tip detection and be utilized as the inputs for robotic endoscopy.